\documentclass{Interspeech}
\usepackage{devanagari}
\interspeechcameraready

\title{CueBuddy: helping non-native English speakers navigate English-centric STEM education}

\author[affiliation={1}]{Pranav}{Gupta}
\affiliation{Lowe's}{Mooresville, North Carolina}{United States}
\email{pranav.gupta@lowes.com}
\keywords{speech recognition, lexical simplification, scientific keyword detection}

\usepackage{comment}

\begin{document}

\maketitle

\begin{abstract}
    
Students across the world in STEM classes, especially in the Global South, fall behind their peers who are more fluent in English, despite being at par with them in terms of scientific prerequisites. While many of them are able to follow everyday English at ease, key terms in English stay challenging. In most cases, such students have had most of their course prerequisites in a lower resource language. Live speech translation to lower resource languages is a promising area of research, however, models for speech translation can be too expensive on a large scale and often struggle with technical content. In this paper, we describe CueBuddy, which aims to remediate these issues by providing real-time "lexical cues" through technical keyword spotting along real-time multilingual glossary lookup to help students stay up to speed with complex English jargon without disrupting their concentration on the lecture. We also describe the limitations and future extensions of our approach. 
\end{abstract}

\section{Introduction}

English is the dominant language of instruction in STEM education globally, including in countries where it is not the first language of most students. While non-native English speakers often have sufficient conversational fluency, they face substantial barriers in understanding domain-specific scientific vocabulary. This challenge is especially acute for students in the Global South who have completed prerequisite coursework in lower-resource languages \cite{xie2025language}. These linguistic barriers contribute to disproportionate dropout rates and lower performance in STEM disciplines compared to their more privileged peers who experienced greater exposure to scientific education in English through expensive school education or secondary tutoring.

In response, researchers have proposed a range of speech and language technologies to support equitable learning, including real-time translation systems \cite{sika2025dragoman,chaudhari2022speech}, automatic speech recognition \cite{santhanalakshmi2024stt,kulkarni2024realtime}, and speech-to-text pipelines adapted for educational use \cite{dash2024speech}. Classroom implementations of speech translation show promise, but many current systems are costly to scale and exhibit low performance on domain-specific terminology \cite{lewandowski2016translation,zainuddin2023review}.

Moreover, while full-sentence translation can aid general comprehension, it may disrupt the attention and cognitive processing required for engaging with scientific material. Many students lie somewhere between the two extremes of not knowing English and being fluent in all aspects of English. They might know most words in a sentence spoken by the instructor, but  Additionally, such tools currently require expensive, cloud-based processing that might not be practical in offline classroom settings. Educators and linguists have long highlighted the benefits of multilingual glossaries and just-in-time vocabulary support to scaffold understanding without breaking lecture flow \cite{colorin2023glossary,queensland2023glossary}. As language models demonstrate an ever-increasing performance in multilingual tasks, rapid prototyping of such ideas to best suit students' needs becomes more viable, allowing us to explore solutions beyond just real-time translation and transcription. 

To address these challenges, we present \textbf{CueBuddy}, a real-time, lightweight system that delivers lexical support for technical terms during STEM lectures. Instead of relying on full translation or full-text speech recognition, CueBuddy identifies key scientific terms in real time and provides definitions or translations from curated multilingual glossaries. Our approach is designed to be low-resource, scalable, and responsive to the needs of students in both high- and low-bandwidth environments. This paper outlines the design, implementation, and early evaluation of CueBuddy, along with its limitations and directions for future research.

\section{Methodology and Related Work}
The CueBuddy system aims to detect STEM keywords appearing in real-time speech and update their short explanations in the user's desired language by looking up a pre-computed multilingual glossary. 

\begin{figure}[t]
  \centering
  \includegraphics[width=\linewidth]{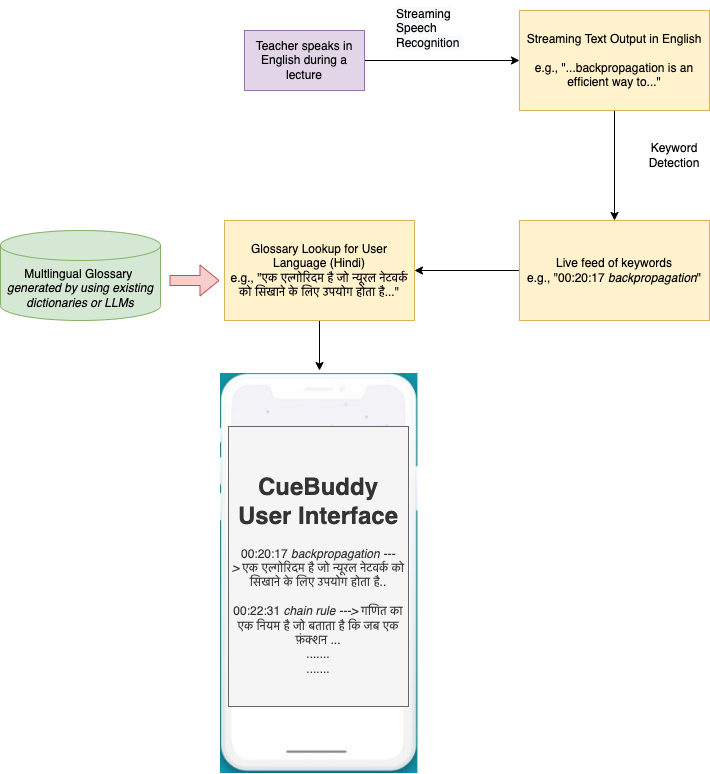}
  \caption{Schematic diagram describing the CueBuddy app}
  \label{fig:schematic}
\end{figure}

\subsection{(Streaming) speech recognition}
The crucial and most challenging aspect of this flow is detecting speech in real time with low latency. Models such as Whisper-tiny \cite{pmlr-v202-radford23a} show promise in this area, and finetuning them on scientific English speech-to-text datasets such as BhasaAnuvaad \cite{jain2024bhasaanuvaadspeechtranslationdataset} can boost their domain-specific performance. Challenges include noisy environments, multiple speakers, underrepresented accents of English, and new vocabulary not being present in the training data. Methods such as biased decoding \cite{nguyen-etal-2024-improving} can improve the domain-specific accuracy of the speech-to-text models. We assume that the instruction speaks in English and does not code-switch with other languages familiar to the student, because teachers flexible with code-switching typically are more aware of student needs and would explain scientific keywords in the student's language. 
\subsection{Keyword detection}
The next step is keyword detection. From the transcribed text, a keyword extraction model (keyBERT) extracts the relevant keywords, such as ``neural network,'' and ``backpropagation.'' However, the performance of such models might be worse on ASR-generated text compared to human-like text. Also, scientific English involves unusual words that do not resemble everyday English. Further training on synthetic and human-annotated keyword detection datasets could help in addressing this issue. 

Another direction of work could be direct decoding of speech to keywords, which could result in a significant reduction in overall latency, because it merges the keyword detection step with the speech recognition step. 

\subsection{Glossary Lookup}
While students have access to several dictionary apps that translate English words to their native language, such tools might not have a complete coverage of scientific vocabulary. For this, we can use LLMs to generate short, multilingual explanations of scientific terms in English. For example, for a student whose native language is Hindi, the glossary will look up the corresponding Hindi translation. Given the domain-specific nature of STEM terms and a lack of sufficient STEM training data in many languages, the glossary might suffer from quality issues. 
\begin{table}[th]
  \caption{Example glossary entry}
  \label{tab:example}
  \centering
  \begin{tabular}{|p{2cm}|p{2cm}|p{2cm}|}
    \toprule
 Term & Hindi & Swahili\\
    \midrule
   backpropagation & ek algorithm hai jo ... (transliterated)& ni mbinu ya kufundisha ...\\
    % $1$                       & $/10$ & $-20$~~~             \\
    % $1$                       & $/1$  & $0$~~~               \\
    % $2$                       & $/1$  & $\approx 6$~~~       \\
    % $3.16$                    & $/1$  & $10$~~~              \\
    % $10$                      & $/1$  & $20$~~~              \\
    \bottomrule
  \end{tabular}
  
\end{table}

% Equations should be placed on separate lines and numbered. We define
% % 
% \begin{align}
%   x(t) &= s(t') \nonumber \\ 
%        &= s(f_\omega(t))
% \end{align}
% % 
% where \(f_\omega(t)\) is a special warping function. Equation \ref{equation:eq2} is a little more complicated.
% % 
% \begin{align}
%   f_\omega(t) &= \frac{1}{2 \pi j} \oint_C 
%   \frac{\nu^{-1k} \mathrm{d} \nu}
%   {(1-\beta\nu^{-1})(\nu^{-1}-\beta)}
%   \label{equation:eq2}
% \end{align}
% % 

\section{Discussion and Future Work}
For CueBuddy to benefit its target user base, extensive experimentation and evaluation is necessary for the models used in the pipeline. We also need to create subject-specific evaluations. We expect that popular subjects such as math at the first-year university level would have more accuracy, whereas uncommon subjects within engineering or medicine would have less accuracy. 

Apart from evaluations, user experience studies and impact studies for evaluating the effective of the application in a classroom setting is also necessary.

\bibliographystyle{IEEEtran}
\bibliography{paper}

\begin{thebibliography}{10}
\providecommand{\url}[1]{#1}
\csname url@samestyle\endcsname
\providecommand{\newblock}{\relax}
\providecommand{\bibinfo}[2]{#2}
\providecommand{\BIBentrySTDinterwordspacing}{\spaceskip=0pt\relax}
\providecommand{\BIBentryALTinterwordstretchfactor}{4}
\providecommand{\BIBentryALTinterwordspacing}{\spaceskip=\fontdimen2\font plus
\BIBentryALTinterwordstretchfactor\fontdimen3\font minus \fontdimen4\font\relax}
\providecommand{\BIBforeignlanguage}[2]{{%
\expandafter\ifx\csname l@#1\endcsname\relax
\typeout{** WARNING: IEEEtran.bst: No hyphenation pattern has been}%
\typeout{** loaded for the language `#1'. Using the pattern for}%
\typeout{** the default language instead.}%
\else
\language=\csname l@#1\endcsname
\fi
#2}}
\providecommand{\BIBdecl}{\relax}
\BIBdecl

\bibitem{xie2025language}
Y.~Xie and H.~Jiao, ``Enhancing language education in developing countries through intelligent transformation: A comprehensive study,'' \emph{Educational Technology Research and Development}, 2025.

\bibitem{sika2025dragoman}
K.~Sika, J.~Chandru, G.~Madhappan, and M.~Vikram, ``Dragoman ai: Real-time speech translation for educational settings,'' \emph{International Journal of Scientific and Applied Technology}, 2025.

\bibitem{chaudhari2022speech}
\BIBentryALTinterwordspacing
S.~Chaudhari, A.~Shukla, and T.~Gaware, ``Real-time direct speech-to-speech translation,'' \emph{International Research Journal of Engineering and Technology (IRJET)}, vol.~9, no.~1, 2022. [Online]. Available: \url{https://www.irjet.net/archives/V9/i1/IRJET-V9I1104.pdf}
\BIBentrySTDinterwordspacing

\bibitem{santhanalakshmi2024stt}
\BIBentryALTinterwordspacing
K.~Santhanalakshmi, K.~Gunal, and D.~Mohan~Raj, ``Speech-to-text and language translation system,'' \emph{International Journal of Research Publication and Reviews}, 2024. [Online]. Available: \url{https://ijrpr.com/uploads/V5ISSUE12/IJRPR36065.pdf}
\BIBentrySTDinterwordspacing

\bibitem{kulkarni2024realtime}
\BIBentryALTinterwordspacing
I.~Kulkarni, S.~Saluja, and V.~Awari, ``Real-time language translator,'' \emph{International Journal of Engineering Research \& Technology (IJERT)}, 2024. [Online]. Available: \url{https://www.ijert.org/real-time-language-translator-2}
\BIBentrySTDinterwordspacing

\bibitem{dash2024speech}
A.~Dash, ``Speech technology and machine learning for inclusive education in developing regions,'' \emph{Journal of Educational Technology}, 2024.

\bibitem{lewandowski2016translation}
N.~Lewandowski, ``Automatic speech translation in the classroom and lecture setting,'' in \emph{New Trends in Audiovisual Translation}.\hskip 1em plus 0.5em minus 0.4em\relax John Benjamins, 2016.

\bibitem{zainuddin2023review}
\BIBentryALTinterwordspacing
N.~Zainuddin, ``Technology enhanced language learning research trends and practices: A systematic review (2020–2022),'' \emph{The Electronic Journal of e-Learning}, 2023. [Online]. Available: \url{https://files.eric.ed.gov/fulltext/EJ1388518.pdf}
\BIBentrySTDinterwordspacing

\bibitem{colorin2023glossary}
\BIBentryALTinterwordspacing
{Color{\'i}n Colorado}, ``Multilingual glossaries: A research-based strategy for english language learners,'' 2023. [Online]. Available: \url{https://www.colorincolorado.org/teaching-ells/ell-classroom-strategy-library/multilingual-glossaries}
\BIBentrySTDinterwordspacing

\bibitem{queensland2023glossary}
\BIBentryALTinterwordspacing
{Queensland Department of Education}, ``Development of a multilingual glossary of school-based terminology,'' 2023. [Online]. Available: \url{https://education.qld.gov.au/about/reporting-data-research/research/Documents/multilingual-glossary-school-based-terminology.pdf}
\BIBentrySTDinterwordspacing

\bibitem{pmlr-v202-radford23a}
\BIBentryALTinterwordspacing
A.~Radford, J.~W. Kim, T.~Xu, G.~Brockman, C.~Mcleavey, and I.~Sutskever, ``Robust speech recognition via large-scale weak supervision,'' in \emph{Proceedings of the 40th International Conference on Machine Learning}, ser. Proceedings of Machine Learning Research, A.~Krause, E.~Brunskill, K.~Cho, B.~Engelhardt, S.~Sabato, and J.~Scarlett, Eds., vol. 202.\hskip 1em plus 0.5em minus 0.4em\relax PMLR, 23--29 Jul 2023, pp. 28\,492--28\,518. [Online]. Available: \url{https://proceedings.mlr.press/v202/radford23a.html}
\BIBentrySTDinterwordspacing

\bibitem{jain2024bhasaanuvaadspeechtranslationdataset}
\BIBentryALTinterwordspacing
S.~Jain, A.~Sankar, D.~Choudhary, D.~Suman, N.~Narasimhan, M.~S. U.~R. Khan, A.~Kunchukuttan, M.~M. Khapra, and R.~Dabre, ``Bhasaanuvaad: A speech translation dataset for 13 indian languages,'' 2024. [Online]. Available: \url{https://arxiv.org/abs/2411.04699}
\BIBentrySTDinterwordspacing

\bibitem{nguyen-etal-2024-improving}
\BIBentryALTinterwordspacing
M.-T. Nguyen, D.~P. Nguyen, T.-H. Luu, X.-Q. Nguyen, T.-D. Nguyen, and J.~Yang, ``Improving speech recognition with jargon injection,'' in \emph{Proceedings of the 25th Annual Meeting of the Special Interest Group on Discourse and Dialogue}, T.~Kawahara, V.~Demberg, S.~Ultes, K.~Inoue, S.~Mehri, D.~Howcroft, and K.~Komatani, Eds.\hskip 1em plus 0.5em minus 0.4em\relax Kyoto, Japan: Association for Computational Linguistics, Sep. 2024, pp. 490--499. [Online]. Available: \url{https://aclanthology.org/2024.sigdial-1.42/}
\BIBentrySTDinterwordspacing

\end{thebibliography}

\end{document}